\documentclass[journal]{IEEEtran}

\ifCLASSINFOpdf
\else
   \usepackage[dvips]{graphicx}
\fi
\usepackage{url}

\usepackage{graphicx}

\begin{document}

\title{Gated Recurrent Unit for Video Denoising}

\author{Kai Guo, \IEEEmembership{Member, IEEE}, Seungwon Choi, Jongseong Choi
\thanks{This work has been submitted to the IEEE for possible publication. Copyright may be transferred without notice, after which this version may no longer be accessible.}
\thanks{K. Guo, S. Choi and J. Choi are with Samsung Electronics, Suwon-si, 16677, South Korea (e-mail: visionkai@gmail.com; sw5190.choi@samsung.com; jongseong.choi@samsung.com).}}

\markboth{Journal of \LaTeX\ Class Files, Vol. 14, No. 8, August 2015}
{Shell \MakeLowercase{\textit{et al.}}: Bare Demo of IEEEtran.cls for IEEE Journals}
\maketitle

\begin{abstract}
Current video denoising methods perform temporal fusion by designing convolutional neural networks (CNN)
 or combine spatial denoising with temporal fusion into basic recurrent neural networks (RNNs).
However, there have not yet been works which adapt gated recurrent unit (GRU) mechanisms for video denoising.
In this letter, we propose a new video denoising model based on GRU, namely GRU-VD. 
First, the reset gate is employed to mark the content related to the current frame in the previous frame output.
Then the hidden activation works as an initial spatial-temporal denoising with the help from the marked relevant content.
Finally, the update gate recursively fuses the initial denoised result with previous frame output to further increase accuracy.
To handle various light conditions adaptively, the noise standard deviation of the current frame is also fed to these three modules.
A weighted loss is adopted to regulate initial denoising and final fusion at the same time.
The experimental results show that the GRU-VD network not only can achieve better quality than state of the arts objectively and subjectively, but also can obtain satisfied subjective quality on real video.
\end{abstract}


\IEEEpeerreviewmaketitle

\vspace{-0.1cm}

\section{Introduction}


\IEEEPARstart{A}{lthough} photographic sensors have made immense progress, various disturbing factors deteriorate the image quality \cite{Foi08:TIP, Hasinoff10:CVPR}, such as shot and readout noise. 
For this reason, denoising is an essential step in image signal processing (ISP) to enhance quality.
Currently convolutional neural networks (CNN) based methods have dominated the state of the arts.
Video denoising explores temporal coherence, hence achieves better quality than single image denoising.

In recent years, recurrent neural networks (RNNs) \cite{Rumelhart86:Nature} including long short-term memory (LSTM) \cite{Hochreiter97:NC} and gated recurrent unit (GRU) \cite{Cho14:SSST} have achieved state of the art in many temporal applications, 
such as speech recognition \cite{Hinton12:SPM} and machine translation \cite{Bahdanau15:ICLR}. 
Current video denoising works perform temporal fusion by designing CNN \cite{Claus19:CVPRW, Tassano19:ICIP, Tassano20:CVPR, Yue20:CVPR, Wang19:CVPRW}
or combine spatial denoising with temporal fusion into basic RNNs \cite{Fuoli19:ICCVW, Godard18:ECCV, Maggioni21:CVPR}.
However, there have not yet been works which adapt GRU with gate mechanisms for video denoising.

\begin{figure}[t]

    \begin{minipage}[b]{1.0\linewidth}
      \centerline{\includegraphics[width= 3.6in]{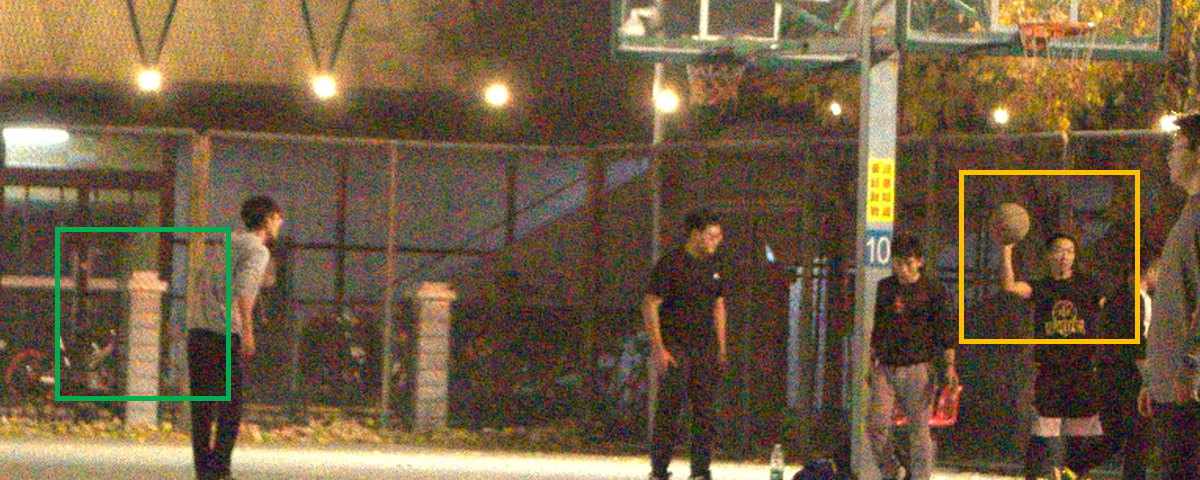}}
      \vspace{-0.05cm}
      \centerline{(a)}\medskip
    \end{minipage}

    \begin{minipage}[b]{0.31\linewidth}
      \centerline{\includegraphics[width= 1.12in]{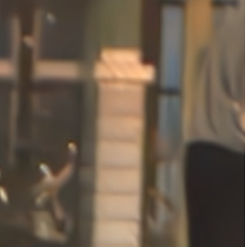}}
      \vspace{-0.05cm}
      \centerline{(b)}\medskip
    \end{minipage}
    \hfill
    \begin{minipage}[b]{0.32\linewidth}
      \centerline{\includegraphics[width= 1.13in]{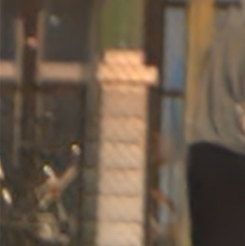}}
      \vspace{-0.05cm}
      \centerline{(c)}\medskip
    \end{minipage}
    \hfill
    \begin{minipage}[b]{0.32\linewidth}
      \centerline{\includegraphics[width= 1.2in]{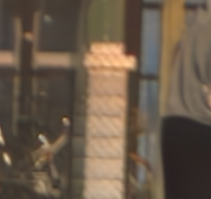}}
      \vspace{-0.05cm}
      \centerline{(d)}\medskip
    \end{minipage}

    \begin{minipage}[b]{0.31\linewidth}
      \centerline{\includegraphics[width= 1.12in]{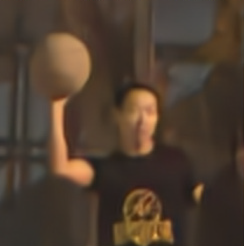}}
      \vspace{-0.05cm}
      \centerline{(e)}\medskip
    \end{minipage}
    \hfill
    \begin{minipage}[b]{0.32\linewidth}
      \centerline{\includegraphics[width= 1.13in]{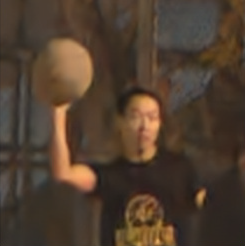}}
      \vspace{-0.05cm}
      \centerline{(f)}\medskip
    \end{minipage}
    \hfill
    \begin{minipage}[b]{0.32\linewidth}
      \centerline{\includegraphics[width= 1.2in]{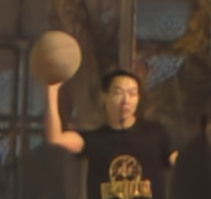}}
      \vspace{-0.05cm}
      \centerline{(g)}\medskip
    \end{minipage}

\vspace{-0.3cm}

    \caption {The proposed GRU-VD has better denoising and detail preservation than the state-of-the-art RViDeNet \cite{Yue20:CVPR} and EMVD \cite{Maggioni21:CVPR}. (a) Input noisy frame; (b, e) RViDeNet; (c, f)  EMVD; (d, g) GRU-VD.} \label{fig:expe_scene4}

\vspace{-0.4cm}
    \end{figure}


In this letter, we propose a new video denoising model based on GRU, namely GRU-VD, to use gate mechanisms for exploring temporal coherence efficiently.
More specifically, first the reset gate is employed to mark the content related to the current frame in the previous frame output.
Secondly, the hidden activation performs initial spatial-temporal denoising with the help of marked relevant content.
Finally, the update gate recursively fuses the initial denoising result with previous frame output to further improve accuracy.
The noise standard deviation of the current frame is also fed to these three modules to deal with various light conditions.
A weighted sum loss is adopted to regulate initial denoising and final fusion simultaneously.
The experimental results show that the GRU-VD network not only can achieve better quality than state of the arts objectively and subjectively, but also can obtain satisfied subjective quality on real video.
A subjective comparison example is shown in Fig. \ref{fig:expe_scene4}, we can see the superiority of the proposed method.
The major contributions of this work are summarized as follows:

    \begin{itemize}

      \item
       GRU-VD is the world-first GRU based video denoising network, which employs gate mechanisms to explore temporal coherence efficiently. 
       In contrast to GRU, GRU-VD employs ReLU instead of Tanh function for hidden activation, to avoid the negative denoising value;
       In addition, it uses the initial denoising, the relevance weight of reset gate and the previous frame output as input for update gate, to improve the fusion accuracy. 


      \item
       GRU-VD provides a new network architecture for video denoising, each module has clear physical meaning. 
       It performs the initial denoising spatially and temporally at the same time, then further improves the accuracy by temporal fusion with the previous frame result. 

    \end{itemize}

\vspace{-0.1cm}

\section{Related works}

Because there are redundant similar textures in the natural images, the classical image denoising methods explore patch aggregation, such as the non-local algorithm \cite{Buades05:CVPR} and BM3D method \cite{Dabov07:TIP}.
At present, the CNN based methods dominate state of the arts both objectively and subjectively.
They generally contain several convolution layers with skip connections and non-linear activation functions, 
such as recursively branched deconvolutional network RBDN \cite{Santhanam17:CVPR}, multi-level wavelet based network MWCNN \cite{Liu18:CVPR}, feed-forward blind denoising network DnCNN \cite{Zhang17:TIP}, fast and flexible denoising network FFDNet \cite{Zhang18:TIP} and residual spatial-adaptive denoising network SADNet \cite{Chang20:ECCV}.

Generally, video denoising methods explore to aggregate temporal coherent information, 
and can be grouped technically into explicit- and implicit-motion methods. 
The explicit-motion methods use patch matching \cite{Davy19:ICIP, Maggioni12:TIP}, optical flow \cite{Sajjadi18:CVPR} or kernel prediction \cite{Vogels18:ACMGraphics, Mildenhall18:CVPR} to obtain motion information, 
and align frames for temporal aggregation. 
However, motion estimation is a challenging task, its error can be amplified by the subsequent aggregation.
Therefore, the implicit-motion methods are proposed, wherein the motion estimation and frames alignment are implicitly processed inside temporal fusion. 
Specifically, ViDeNN \cite{Claus19:CVPRW}, fastDVDnet \cite{Tassano19:ICIP, Tassano20:CVPR}, RViDeNet \cite{Yue20:CVPR} and EDVR \cite{Wang19:CVPRW} sequentially apply spatial denoising and temporal fusion with assistants from pyramid alignment, spatial and temporal attention etc.
Based on this, RLSP \cite{Fuoli19:ICCVW}, MFSR \cite{Godard18:ECCV} and EMVD \cite{Maggioni21:CVPR} employ the recurrent scheme of basic RNNs, wherein the image features from the previous frame are employed as additional input.

\begin{figure}[t]

    \begin{minipage}[b]{0.49\linewidth}
      \centerline{\includegraphics[width= 1.8in]{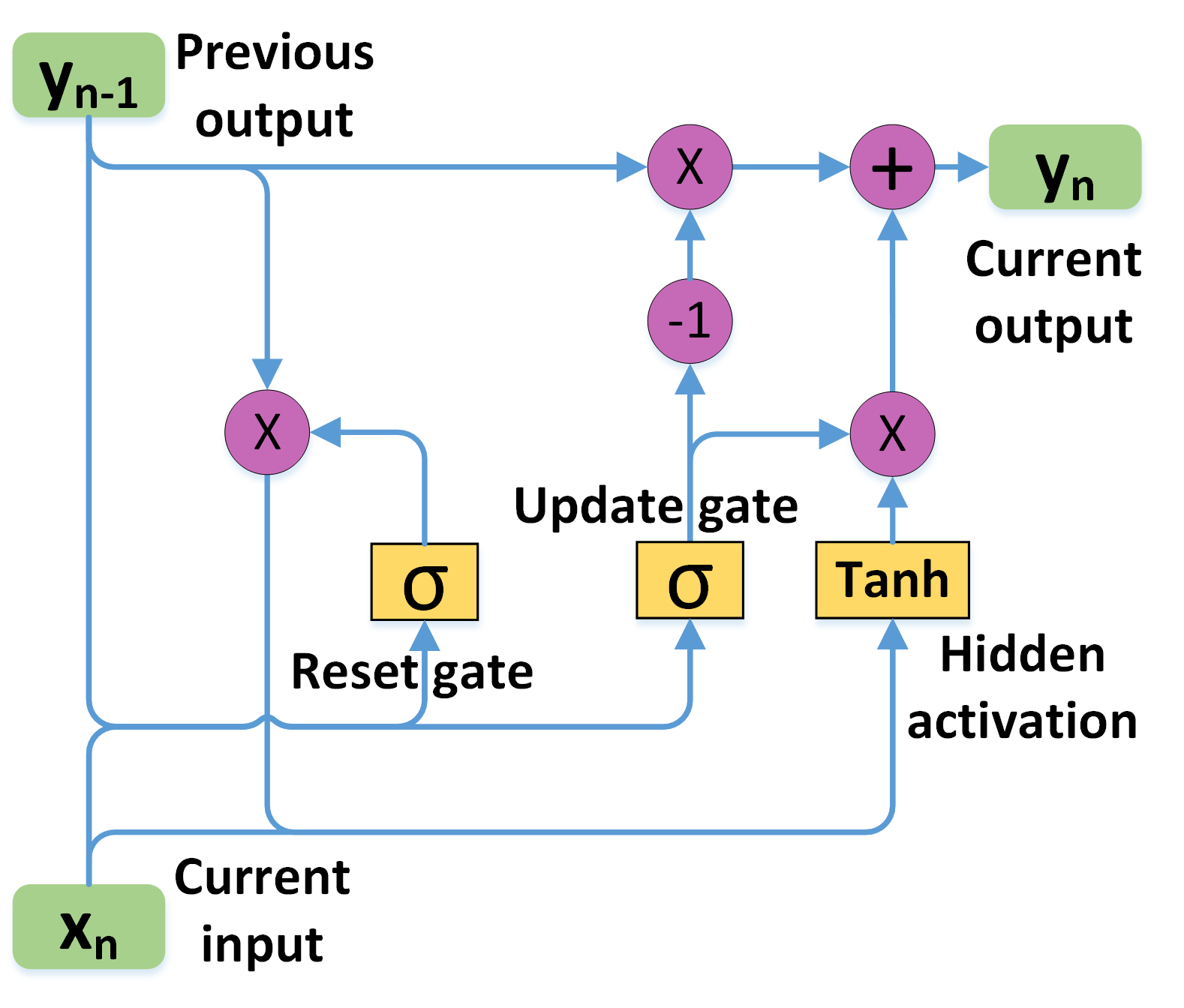}}
      \vspace{0.00cm}
    \end{minipage}
    \hfill
    \begin{minipage}[b]{0.49\linewidth}
      \centerline{\includegraphics[width= 1.8in]{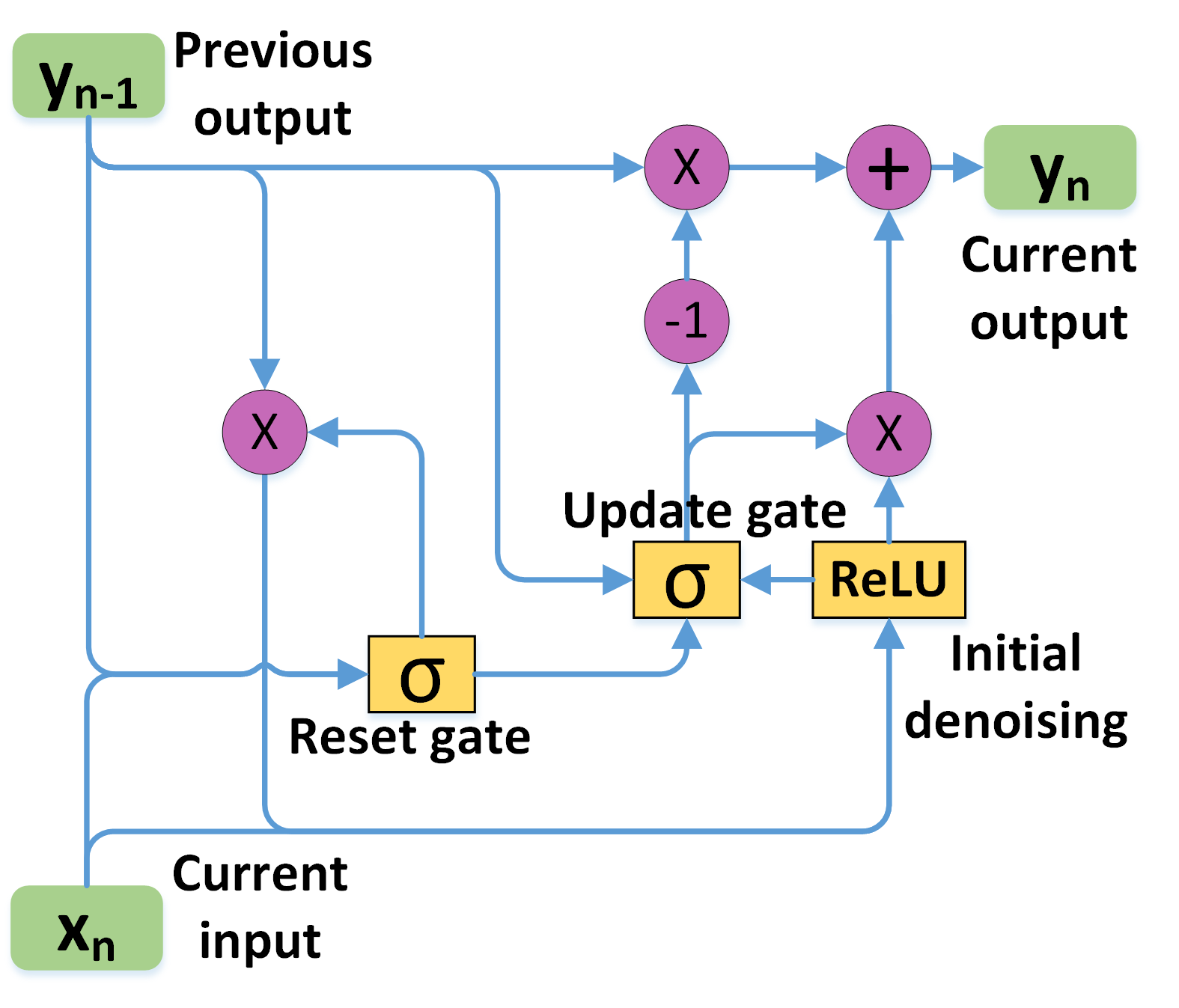}}
      \vspace{0.00cm}
    \end{minipage}

\vspace{-0.2cm}

    \caption {Graphical illustrations of the standard GRU (left) and the proposed GRU-VD (right). "$\times$" and "+" are element-wise multiplication and summation, respectively. "-1" means subtracted by one. "$\sigma$", "Tanh" and "ReLU" are CNN networks with sigmoid, tangent and ReLU activations, respectively.} \label{fig:graph_GRU_VD}
\vspace{-0.4cm}
    \end{figure}

\vspace{-0.1cm}

\section{GRU-VD network}

\subsection{Noise Model}

The video noise is mainly derived from shot and readout noise \cite{Brooks19:CVPR}. The shot noise comes from photon arrival statistics, and can be modeled by a Poisson process whose mean value is the true light intensity. 
The readout noise is caused by imprecision in readout circuitry, and it is modeled by a Gaussian distribution. These two noise can be approximated by a heteroscedastic Gaussian function $\eta$ \cite{Foi08:TIP}, its variance is written as:
    \begin{equation}
    \label{eq:noise_var}
    v_n(\eta_n) = a_ny_n+b_n
    \end{equation}
where $n$ represents the frame index of video, $y$ is the noise-free clean frame. $a_n$ and $b_n$ are the parameters of shot and readout noise, respectively, they are determined by the digital gains and analog of the sensor.
Then the noise observation model can be written as:
    \begin{equation}
    \label{eq:noise_model}
    x_n(i) = y_n(i)+\eta_n(i)
    \end{equation}
where $i$ is the spatial location of the pixel, $x$ is the observed noisy frame. 
Because there are many robust noise estimation methods, we assume the noise parameters are known.

\begin{figure*}
\centerline{\includegraphics[width=6.0in]{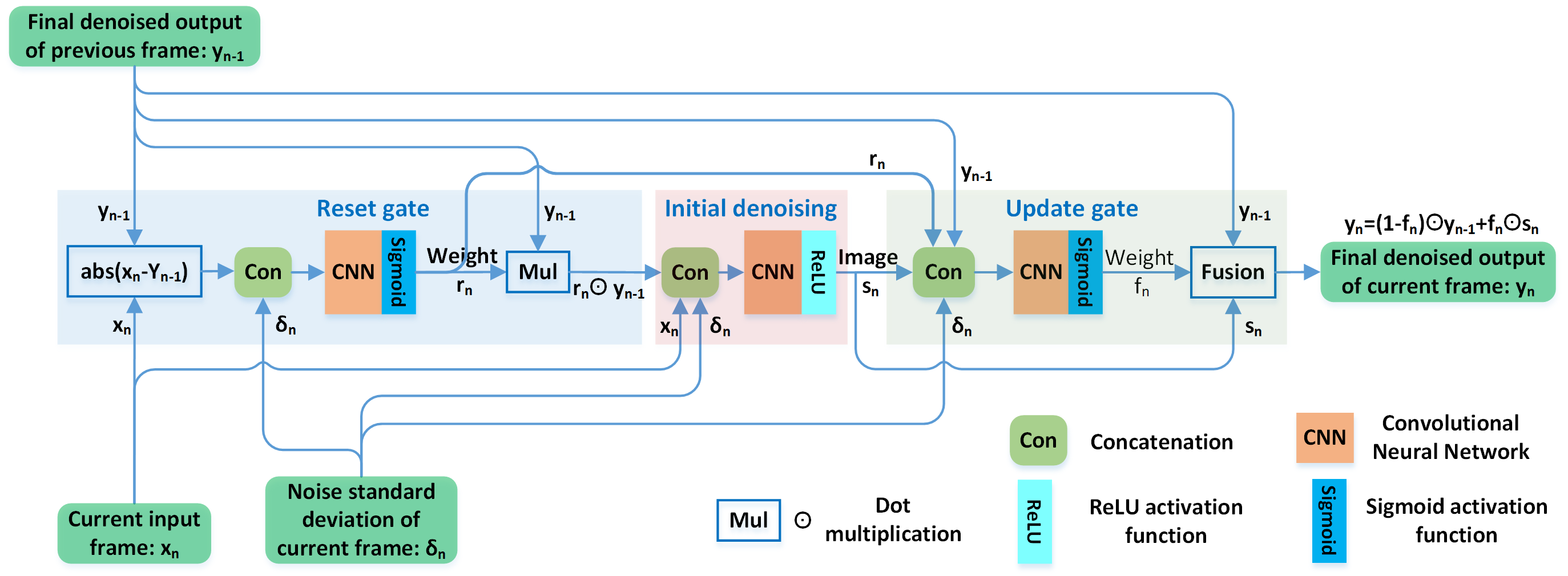}}
\vspace{-0.2cm}
\caption{The detailed network architecture of proposed GRU-VD.}
\label{fig:flowchart_GRU_VD}
\vspace{-0.4cm}
\end{figure*}

\vspace{-0.2cm}

\subsection{GRU Network}

GRU is a kind of RNNs with gating mechanisms, it is like LSTM but with fewer parameters. The architecture of GRU is shown in Fig. \ref{fig:graph_GRU_VD} (left).
The current input frame $x_n$ and the previous frame output $y_{n-1}$ are employed as input. 
There is a reset gate $r$ to identify the useful content of $y_{n-1}$, 
a hidden activation $s$ as preprocessing to get a candidate,
 and an update gate $f$ to estimate the final fusion output.
All the equations of GRU can be written as:
    \begin{equation}
    \label{eq:GRU_equations}
    \begin{array}{rl}
            & r_n = \sigma(W_{r}x_{n} + U_{r}y_{n-1} + b_{r})   \\
            & s_n = tanh(W_{s}x_{n} + U_{s}(r_{n}{\odot}y_{n-1}) + b_{s})   \\
            & f_n = \sigma(W_{f}x_{n} + U_{f}y_{n-1} + b_{f})   \\
            & y_n = (1-f_{n}){\odot}y_{n-1} + f_{n}{\odot}s_{n}
    \end{array}
    \end{equation}
where $n$ is the frame index, matrices $W_r$, $W_s$, $W_f$, $U_r$, $U_s$, $U_f$ and vectors $b_r$, $b_s$, $b_f$ are the parameters of model.
 $\sigma$ and $tanh$ represent element-wise sigmoid and hyperbolic tangent functions, respectively. $\odot$ represents element-wise multiplication.

\vspace{-0.2cm}

\subsection{Proposed GRU-VD Network}

To achieve better quality, the spatial denoising and temporal fusion need to be combined more efficiently. 
The relevant content from previous frame output can assist spatial denoising of the current frame.
The temporal fusion between initial denoising and previous frame output will improve the accuracy further, 
and the fusion weight can be inferred from the initial denoising, the relevant weight as well as the previous frame output.
To realize these observations, we propose GRU-VD network based on GRU, as shown in Fig. \ref{fig:graph_GRU_VD} (right).
The major characteristics of the proposed GRU-VD are as follows:

    \begin{itemize}

      \item 
       The reset gate with sigmoid activation detects the relevance weight of previous frame result. 
       Multiplying the previous frame result with this weight can mark its relevant content.

      \item
       Based on the marked relevant content, the hidden activation works as an initial spatial-temporal denoising.
       It uses ReLU instead of Tanh activation, so as to avoid the negative denoising value. 

      \item
       The update gate with sigmoid activation predicts the temporal fusion weight, which is used to fuse the initial denoising with the previous frame output.
       It employs the initial denoising result, the relevance weight and the previous frame output as input.

      \item
      To deal with various light conditions, the noise standard deviation of current frame works as an additional input to reset gate, initial denoising and update gate.
      A weighted sum loss is adopted to regulate initial denoising and final fusion at the same time.

    \end{itemize}

More specifically (as shown in Fig. \ref{fig:flowchart_GRU_VD}), the reset gate $r$ predicts a relevance weight $r_n$, which marks the relevant content to the current frame $x_n$ in the previous frame output $y_{n-1}$.
The noise standard deviation $\delta_n$ of $x_n$, the absolute difference between $y_{n-1}$ and $x_n$ are employed as input:
    \begin{equation}
    \label{eq:resetGate_input}
    Con(\delta_n, |x_n - y_{n-1}|)
    \end{equation}
where $Con$ means concatenation operation along the channel dimension. 
With the help of a CNN and final sigmoid activation, the reset gate outputs a relevance weight matrix $r_n$. 
$y_{n-1}$ is multiplied with $r_n$ to mark its relevant content.

With respect to the initial denoising, the marked relevant content $r_n {\odot} y_{n-1}$, the current frame $x_n$ and its noise standard deviation $\delta_n$ are employed as input:
    \begin{equation}
    \label{eq:spatialDenoising_input}
    Con(r_n {\odot} y_{n-1}, x_n, \delta_n)
    \end{equation}

After processed by a CNN and final ReLU activation, an initial spatial-temporal denoised frame $s_n$ is obtained. As aforementioned, the ReLU activation is employed to avoid negative value.

For temporal fusion, the update gate $f$ employs the initial denoised frame $s_n$, the previous frame result $y_{n-1}$, the relevance weight $r_n$ and the noise standard deviation $\delta_n$ as input:
    \begin{equation}
    \label{eq:updateGate_input}
    Con(s_n, y_{n-1}, r_n, \delta_n)
    \end{equation}

The update gate with a CNN and final sigmoid activation predicts a fusion weight $f_n$, which is used to weighted average the previous frame output $y_{n-1}$ and the initial denoised frame $s_n$:
    \begin{equation}
    \label{eq:final_fusion}
       y_n = (1-f_n){\odot}y_{n-1} + f_n{\odot}s_n
    \end{equation}

All the equations of GRU-VD can be summarized as follows:
    \begin{equation}
    \label{eq:GRU_VD_equations}
    \begin{array}{rl}
            & r_n = \sigma(W_{r}|x_n - y_{n-1}| + V_{r}{\delta_n} + b_{r})   \\
            & s_n = ReLU(W_{s}x_{n} + U_{s}(r_{n}{\odot}y_{n-1}) + V_{s}{\delta_n} + b_{s})   \\
            & f_n = \sigma(W_{f}s_{n} + U_{f}y_{n-1} + V_{f}{\delta_n} + T_{f}{r_n} + b_{f})   \\
            & y_n = (1-f_{n}){\odot}y_{n-1} + f_{n}{\odot}s_{n}
    \end{array}
    \end{equation}
where matrices $W_r$, $W_s$, $W_f$, $U_s$, $U_f$, $V_r$, $V_s$, $V_f$, $T_f$ and vectors $b_r$, $b_s$, $b_f$ are the parameters of model.
 $\sigma$ and $ReLU$ represent element-wise sigmoid and ReLU activations, respectively. $\odot$ represents element-wise multiplication.


   \begin{table}[t]
	\caption{Average PSNR/SSIM comparison on the CRVD dataset. The best and second-best results are highlighted and underlined.}
	\centering
	\scalebox{1}{
		\begin{tabular}{|c|c|c|c|c|}
			\hline
			&\multicolumn{2}{|c|}{raw}&\multicolumn{2}{|c|}{sRGB}\\
			\hline
			Model&PSNR&SSIM&PSNR&SSIM\\
			\hline
			FastDVDnet \cite{Tassano20:CVPR}&44.30&0.9891&39.91&0.9812\\
			RViDeNet \cite{Yue20:CVPR}&44.08&0.9881&40.03&0.9802\\
			EDVR \cite{Wang19:CVPRW}&\underline{44.71}&\underline{0.9902}&\underline{40.89}&\underline{0.9838}\\
			EMVD \cite{Maggioni21:CVPR}&44.05&0.9890&39.53&0.9796\\
			Ours&\textbf{45.06}&\textbf{0.9981}&\textbf{41.14}&\textbf{0.9941}\\
			\hline
		\end{tabular}
	}
	\label{tab:PSNR_SSIM_CRVD}
	\vspace{-0.4cm}
\end{table}

\vspace{-0.2cm}

\subsection{CNN and Loss Function}

With respect to the CNNs of reset gate, initial denoising and update gate, we select information multi-distillation network (IMDN) \cite{Hui19:ACMMM}. 
Because it can extract hierarchical features and aggregate these features according to their importance.
The larger receptive field of CNN is very useful to handle various motions in video denoising. 
Therefore we employ $12$ information multi-distillation blocks (IMDB) for each IMDN.

For GRU-VD, the initial denoising output $s_n$ should be similar with the ground truth $\hat{y}_n$, and the final fusion result $f_n$ after the update gate needs to further improve the accuracy.
Hence we employ a weighted sum of two $L_1$ loss functions to regulate $s_n$ and $f_n$ simultaneously:
    \begin{equation}
    \label{eq:loss_function}
       L = w_{1}|y_n - \hat{y}_n| + w_{2}|s_n - \hat{y}_n|
    \end{equation}
where $w_{1}$ and $w_{2}$ are the weights, they are set as $0.1$ and $1$, respectively.

\vspace{-0.2cm}

\section{Experiments}

In order to validate the effectiveness and robustness of our proposed GRU-VD, we compare it with five state-of-the-art video denoising methods: VBM4D \cite{Maggioni12:TIP}, FastDVDnet \cite{Tassano20:CVPR}, EDVR \cite{Wang19:CVPRW}, RViDeNet \cite{Yue20:CVPR} and EMVD \cite{Maggioni21:CVPR} on a video benchmark dataset \cite{Yue20:CVPR}, then evaluate it on a real-world video.
The benchmark dataset comprises a real raw video dataset (CRVD), which is captured by a SONY IMX385 sensor, and a synthesized dataset (SRVD) \cite{Chen18:CVPR}. 
All these videos contain five different ISO values which range from $1600$ to $25600$.
For fair comparison, we adopt the same training datasets with RViDeNet \cite{Yue20:CVPR} and EMVD \cite{Maggioni21:CVPR}: SRVD dataset and the indoor scenes $1\sim6$ of CRVD dataset. 
The objective and subjective comparisons are performed based on the indoor scenes $7\sim11$ and the outdoor scenes of CRVD dataset, respectively.

\begin{figure} [t]

    \begin{minipage}[b]{1.0\linewidth}
      \centerline{\includegraphics[width= 3.6in]{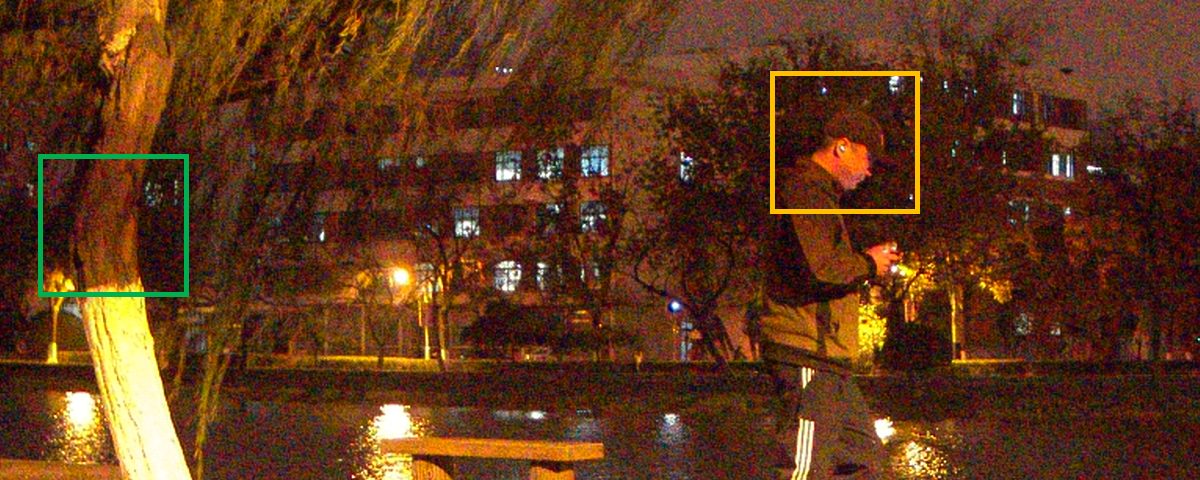}}
      \vspace{-0.05cm}
      \centerline{(a)}\medskip
    \end{minipage}

    \begin{minipage}[b]{0.30\linewidth}
      \centerline{\includegraphics[width= 1.16in]{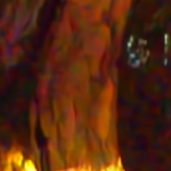}}
      \vspace{-0.05cm}
      \centerline{(b)}\medskip
    \end{minipage}
    \hfill
    \begin{minipage}[b]{0.30\linewidth}
      \centerline{\includegraphics[width= 1.16in]{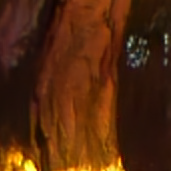}}
      \vspace{-0.05cm}
      \centerline{(c)}\medskip
    \end{minipage}
    \hfill
    \begin{minipage}[b]{0.30\linewidth}
      \centerline{\includegraphics[width= 1.16in]{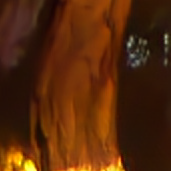}}
      \vspace{-0.05cm}
      \centerline{(d)}\medskip
    \end{minipage}

    \begin{minipage}[b]{0.30\linewidth}
      \centerline{\includegraphics[width= 1.16in]{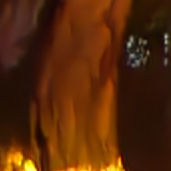}}
      \vspace{-0.05cm}
      \centerline{(e)}\medskip
    \end{minipage}
    \hfill
    \begin{minipage}[b]{0.30\linewidth}
      \centerline{\includegraphics[width= 1.16in]{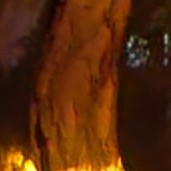}}
      \vspace{-0.05cm}
      \centerline{(f)}\medskip
    \end{minipage}
    \hfill
    \begin{minipage}[b]{0.30\linewidth}
      \centerline{\includegraphics[width= 1.16in]{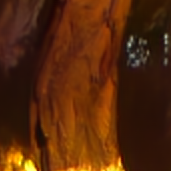}}
      \vspace{-0.05cm}
      \centerline{(g)}\medskip
    \end{minipage}

    \begin{minipage}[b]{0.30\linewidth}
      \centerline{\includegraphics[width= 1.16in]{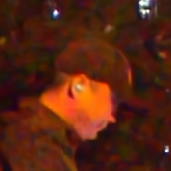}}
      \vspace{-0.05cm}
      \centerline{(h)}\medskip
    \end{minipage}
    \hfill
    \begin{minipage}[b]{0.30\linewidth}
      \centerline{\includegraphics[width= 1.16in]{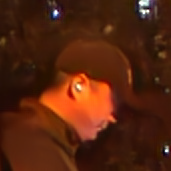}}
      \vspace{-0.05cm}
      \centerline{(i)}\medskip
    \end{minipage}
    \hfill
    \begin{minipage}[b]{0.30\linewidth}
      \centerline{\includegraphics[width= 1.16in]{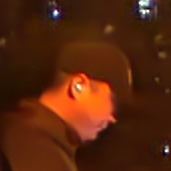}}
      \vspace{-0.05cm}
      \centerline{(j)}\medskip
    \end{minipage}

    \begin{minipage}[b]{0.30\linewidth}
      \centerline{\includegraphics[width= 1.18in]{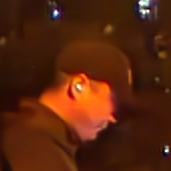}}
      \vspace{-0.05cm}
      \centerline{(k)}\medskip
    \end{minipage}
    \hfill
    \begin{minipage}[b]{0.30\linewidth}
      \centerline{\includegraphics[width= 1.18in]{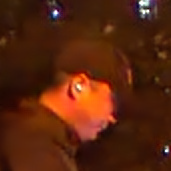}}
      \vspace{-0.05cm}
      \centerline{(l)}\medskip
    \end{minipage}
    \hfill
    \begin{minipage}[b]{0.30\linewidth}
      \centerline{\includegraphics[width= 1.18in]{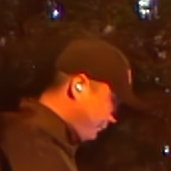}}
      \vspace{-0.05cm}
      \centerline{(m)}\medskip
    \end{minipage}

\vspace{-0.3cm}

    \caption {The proposed GRU-VD exhibits better denoising and detail preservation than the state-of-the-art methods. (a) Input noisy frame; (b, h) VBM4D \cite{Maggioni12:TIP}; (c, i)  FastDVDnet \cite{Tassano20:CVPR}; (d, j) EDVR \cite{Wang19:CVPRW};  (e, k) RViDeNet \cite{Yue20:CVPR};  (f, l) EMVD \cite{Maggioni21:CVPR};  (g, m) GRU-VD.} \label{fig:expe_scene3}
\vspace{-0.4cm}
    \end{figure}

\vspace{-0.3cm}

\subsection{Training}
We set the feature number of each IMDN as $96$, then the total number of GRU-VD parameters is $9M$.
Since our GRU-VD is a recurrent network, its denoising quality depends on the frames number in each training epoch.
Considering fair comparison, we choose the same frames number $25$ as EMVD \cite{Maggioni21:CVPR}.
In each training epoch, $25$ patches with resolution $256\times256$ are randomly cropped from $25$ randomly extracted frames of a video.
The network is trained by adopting Adam optimizer\cite{Kingma15:ICLR} with batch size $16$.
The initial learning rate is set as $1e-4$ and divided by $10$ at every $32000$ epochs. 
We implement our network with the PyTorch framework and train it using a NVIDIA A100 GPU.

\vspace{-0.3cm}

\subsection{Comparison with State-of-the-art Methods}

According to the averaged PSNR and SSIM comparisons in Tab. \ref{tab:PSNR_SSIM_CRVD}, the proposed GRU-VD achieves the best performance. 
The proposed GRU-VD exceeds the PSNR and SSIM values of the second best method EDVR \cite{Wang19:CVPRW} by 0.35 and 0.0079 in raw space, respectively, and by 0.25 and 0.0103 in sRGB space, respectively.
Fig. \ref{fig:expe_scene4} and Fig. \ref{fig:expe_scene3} show subjective comparison examples under low-light scenes with ISO $25600$, wherein the results of other state-of-the-art methods are extracted from the EMVD paper \cite{Maggioni21:CVPR}. 
It is observed that the proposed GRU-VD has obviously reduced noise and recovered details. 
For example, our method recovers better details in the area of trunk and the hat in Fig. \ref{fig:expe_scene3}, and avoids the fake textures and artifacts.

\vspace{-0.3cm}

\subsection{Results on Real Video}

\begin{figure}

    \begin{minipage}[b]{1.0\linewidth}
      \centerline{\includegraphics[width= 3.5in]{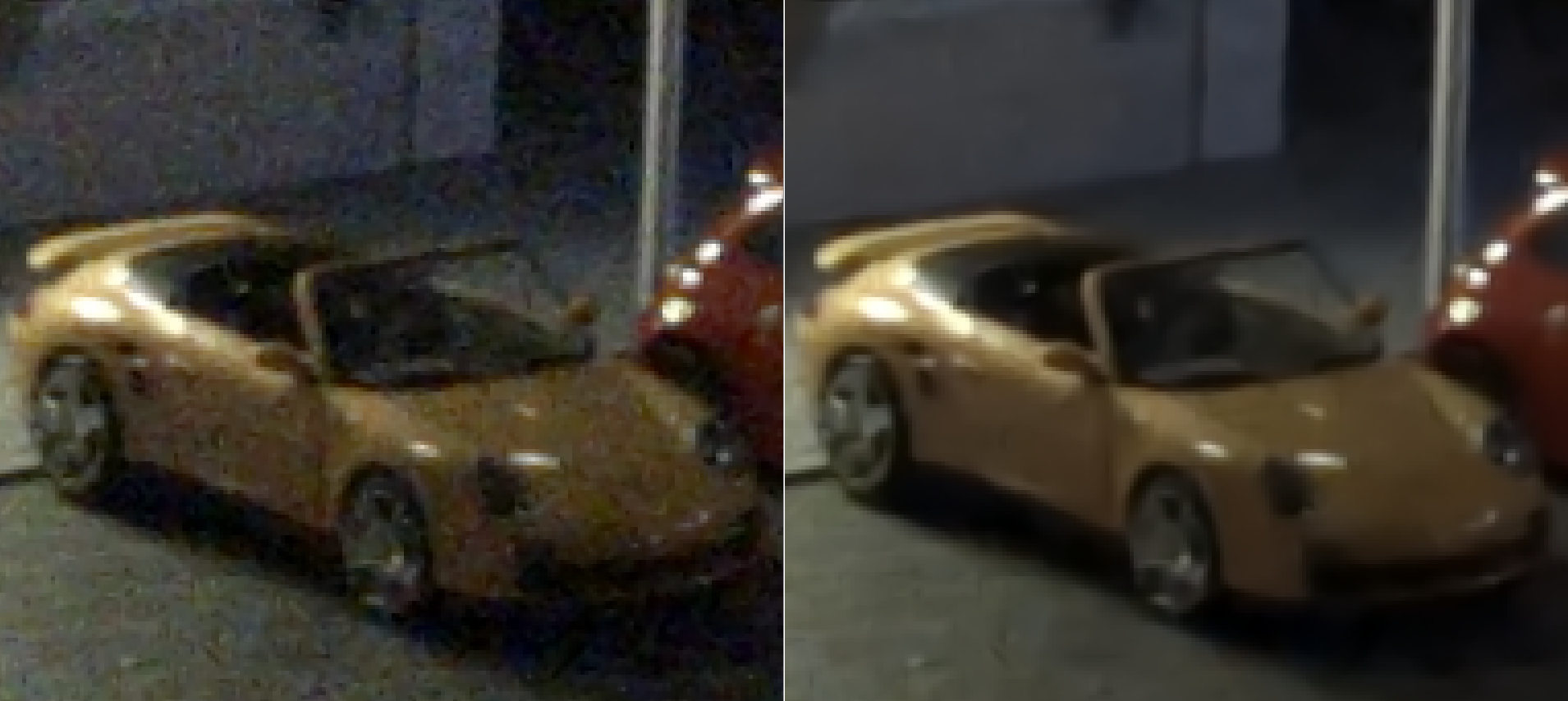}}
      \vspace{0.00cm}
    \end{minipage}

\vspace{-0.2cm}
\caption{Result on real video. Left: input noisy frame. Right: GRU-VD result.}
\label{fig:result_realVideo}
\end{figure}

We test the proposed GRU-VD network on a video captured under low-light conditions by Samsung Galaxy S22 (Samsung ISOCELL HM3 sensor).
To keep the noise of the test video, we disable the denoising module of ISP.
Therefore the test video has serious noise, as shown in Fig. \ref{fig:result_realVideo} (left).
The GRU-VD denoised result is shown in Fig. \ref{fig:result_realVideo} (right). 
We can see that the proposed GRU-VD network not only can reduce the noise efficiently, but also can preserve the image details.

\vspace{-0.2cm}

\section{Conclusion}

In this letter, we have proposed GRU-VD network for video denoising. 
GRU-VD is the world-first video denoising network based on the GRU network, which can efficiently combine spatial and temporal denoising. 
The experimental results show that the GRU-VD network not only can achieve better quality than state of the arts both objectively and subjectively, but also can obtain satisfied subjective denoising quality on real video.
In the future, we will investigate extending the proposed GRU-VD network into other computer vision problems.


\vspace{-0.1cm}


\begin{thebibliography}{31}

\bibitem{Foi08:TIP}
A. Foi, M. Trimeche, V. Katkovnik, and K. Egiazarian,  ``Practical Poissonian-Gaussian Noise Modeling and Fitting for Single-Image Raw-Data,''
  \emph{IEEE Trans. Image Processing}, vol. 17, No. 10, pp. 1737--1754, Sep. 2008.

\bibitem{Hasinoff10:CVPR}
S. Hasinoff, F. Durand, and W. Freeman, ``Noise-optimal capture for high dynamic range photography,''
\emph{IEEE/CVF Conference on Computer Vision and Pattern Recognition (CVPR)}, San Francisco, CA, USA, Jun. 2010, pp. 553--560.

\bibitem{Rumelhart86:Nature}
D. Rumelhart, G. Hinton, and R. Williams,  ``Learning representations by back-propagating errors,''
  \emph{Nature}, vol. 323, No. 6088, pp. 533--536, Oct. 1986.

\bibitem{Hochreiter97:NC}
S. Hochreiter, and J. Schmidhuber,  ``Long Short-Term Memory,''
  \emph{Neural Computation}, vol. 9, No. 8, pp. 1735--1780, Nov. 1997.

\bibitem{Cho14:SSST}
K. Cho, B. Merrienboer, D. Bahdanau, and Y. Bengio, ``On the Properties of Neural Machine Translation: Encoder-Decoder Approaches,''
\emph{Eighth Workshop on Syntax, Semantics and Structure in Statistical Translation (SSST-8)}, Doha, Qatar, Sep. 2014, pp. 103--111.



\bibitem{Hinton12:SPM}
G. Hinton, L. Deng, D. Yu, G. Dahl, A-R. Mohamed, N. Jaitly, A. Senior, V. Vanhoucke, P. Nguyen, T. Sainath, and B. Kingsbury,  ``Deep Neural Networks for Acoustic Modeling in Speech Recognition: The Shared Views of Four Research Groups,''
  \emph{IEEE Signal Processing Magazine}, vol. 29, No. 6, pp. 82--97, Nov. 2012.


\bibitem{Bahdanau15:ICLR}
D. Bahdanau, K. Cho, and Y. Bengio, ``Neural machine translation by jointly learning to align and translate,''
\emph{International Conference on Learning Representations (ICLR)}, San Diego, CA, USA, May. 2015, pp. 1--15.


\bibitem{Claus19:CVPRW}
M. Claus, and J. Gemert, ``ViDeNN: Deep blind video denoising,''
\emph{IEEE/CVF Conference on Computer Vision and Pattern Recognition Workshops (CVPRW)}, Long Beach, CA, USA, Jun. 2019, pp. 1843--1852.


\bibitem{Tassano19:ICIP}
M. Tassano, J. Delon, and T. Veit, ``DVDnet: A fast network for deep video denoising,''
\emph{IEEE International Conference on Image Processing (ICIP)}, Taipei, Taiwan, Sep. 2019, pp. 1805--1809.

\bibitem{Tassano20:CVPR}
M. Tassano, J. Delon, and T. Veit, ``FastDVDnet: Towards Real-Time Deep Video Denoising Without Flow Estimation,''
\emph{IEEE/CVF Conference on Computer Vision and Pattern Recognition (CVPR)}, Seattle, WA, USA, Jun. 2020, pp. 1354--1363.

\bibitem{Yue20:CVPR}
H. Yue, C. Cao, L. Liao, R. Chu, and J. Yang, ``Supervised Raw Video Denoising with a Benchmark Dataset on Dynamic Scenes,''
\emph{IEEE/CVF Conference on Computer Vision and Pattern Recognition (CVPR)}, Seattle, WA, USA, Jun. 2020, pp. 2301--2310.

\bibitem{Wang19:CVPRW}
X. Wang, K. Chan, K. Yu, C. Dong, and C. Loy, ``EDVR: Video Restoration with Enhanced Deformable Convolutional Networks,''
\emph{IEEE/CVF Conference on Computer Vision and Pattern Recognition Workshops (CVPRW)}, Long Beach, CA, USA, Jun. 2019, pp. 1954--1963.

\bibitem{Fuoli19:ICCVW}
D. Fuoli, S. Gu and R. Timofte, ``Efficient Video Super-Resolution through Recurrent Latent Space Propagation,''
\emph{IEEE/CVF International Conference on Computer Vision Workshop (ICCVW)}, Seoul, Korea, Oct. 2019, pp. 3476--3485.

\bibitem{Godard18:ECCV}
C. Godard, K. Matzen, and M. Uyttendaele, ``Deep burst denoising,''
\emph{European Conference on Computer Vision (ECCV)}, Munich, Germany, Sep. 2018, pp. 560--577.

\bibitem{Maggioni21:CVPR}
M. Maggioni, Y. Huang, C. Li, S. Xiao, Z. Fu, and F. Song, ``Efficient Multi-Stage Video Denoising with Recurrent Spatio-Temporal Fusion,''
\emph{IEEE/CVF Conference on Computer Vision and Pattern Recognition (CVPR)}, Nashville, TN, USA, Jun. 2021, pp. 3466--3475.


\bibitem{Maggioni12:TIP}
M. Maggioni, G. Boracchi, A. Foi, and K. Egiazarian,  ``Video Denoising, Deblocking, and Enhancement Through Separable 4-D Nonlocal Spatiotemporal Transforms,''
  \emph{IEEE Trans. Image Processing}, vol. 21, No. 9, pp. 3952--3966, Feb. 2012.

\bibitem{Chen18:CVPR}
C. Chen, Q. Chen, J. Xu, and V. Koltun, ``Learning to See in the Dark,''
\emph{IEEE/CVF Conference on Computer Vision and Pattern Recognition (CVPR)}, Salt Lake City, UT, USA, Jun. 2018, pp. 3291--3300.

\bibitem{Kingma15:ICLR}
D. Kingma, and J. Ba, ``Adam: A Method for Stochastic Optimization,''
\emph{Proceedings of the 3rd International Conference on Learning Representations (ICLR)}, San Diego, CA, USA, May. 2015.


\bibitem{Buades05:CVPR}
A. Buades, B. Coll, and J-M. Morel, ``A non-local algorithm for image denoising,''
\emph{IEEE/CVF Conference on Computer Vision and Pattern Recognition (CVPR)}, San Diego, CA, USA, Jun. 2005, pp. 60--65.


\bibitem{Dabov07:TIP}
K. Dabov, A. Foi, V. Katkovnik, and K. Egiazarian,  ``Image Denoising by Sparse 3-D Transform-Domain Collaborative Filtering,''
  \emph{IEEE Trans. Image Processing}, vol. 16, No. 8, pp. 2080--2095, Jul. 2007.



\bibitem{Santhanam17:CVPR}
V. Santhanam, V.I. Morariu, and L.S. Davis, ``Generalized Deep Image to Image Regression,''
\emph{IEEE/CVF Conference on Computer Vision and Pattern Recognition (CVPR)}, Honolulu, HI, USA, Jun. 2017, pp. 5395--5405.


\bibitem{Liu18:CVPR}
P. Liu, H. Zhang, K. Zhang, L. Lin, and W. Zuo, ``Multi-level wavelet-CNN for image restoration,''
\emph{IEEE/CVF Conference on Computer Vision and Pattern Recognition (CVPR)}, Salt Lake City, UT, USA, Jun. 2018, pp. 886--895.

\bibitem{Zhang17:TIP}
K. Zhang, W. Zuo, Y. Chen, D. Meng, and L. Zhang,  ``Beyond a gaussian denoiser: Residual learning of deep cnn for image denoising,''
  \emph{IEEE Trans. Image Processing}, vol. 26, No. 7, pp. 3142--3155, Jul. 2017.


\bibitem{Zhang18:TIP}
K. Zhang, W. Zuo, and L. Zhang,  ``Ffdnet: Toward a fast and flexible solution for cnn-based image denoising,''
  \emph{IEEE Trans. Image Processing}, vol. 27, No. 9, pp. 4608--4622, Sep. 2018.

\bibitem{Chang20:ECCV}
M. Chang, Q. Li, H. Feng, and Z. Xu, ``Spatial-adaptive network for single image denoising,''
\emph{European Conference on Computer Vision (ECCV)}, Glasgow, United Kingdom, Aug. 2020, pp. 171--187.


\bibitem{Sajjadi18:CVPR}
M. Sajjadi, R. Vemulapalli, and M. Brown, ``Frame-recurrent video super-resolution,''
\emph{IEEE/CVF Conference on Computer Vision and Pattern Recognition (CVPR)}, Salt Lake City, UT, USA, Jun. 2018, pp. 6626--6634.


\bibitem{Davy19:ICIP}
A. Davy, T. Ehret, J-M. Morel, P. Arias, and G. Facciolo, ``A non-local CNN for video denoising,''
\emph{IEEE International Conference on Image Processing (ICIP)}, Taipei, Taiwan, Sep. 2019, pp. 2409--2413.


\bibitem{Vogels18:ACMGraphics}
T. Vogels, F. Rousselle, B. Mcwilliamsl, G. Rothlin, A. Harvill, D. Adler, M. Meyer, and J. Novak, ``Denoising with kernel prediction and asymmetric loss functions,''
\emph{ACM Transactions on Graphics}, vol. 37, No. 4, pp. 1--15, Aug. 2018.


\bibitem{Mildenhall18:CVPR}
B. Mildenhall, J. Barron, J. Chen, D. Sharlet, R.  Ng, and R. Carroll, ``Burst denoising with kernel prediction networks,''
\emph{IEEE/CVF Conference on Computer Vision and Pattern Recognition (CVPR)}, Salt Lake City, UT, USA, Jun. 2018, pp. 2502--2510.





\bibitem{Brooks19:CVPR}
T. Brooks, B. Mildenhall, T. Xue, J. Chen, D. Sharlet, and J. Barron, ``Unprocessing Images for Learned Raw Denoising,''
\emph{IEEE/CVF Conference on Computer Vision and Pattern Recognition (CVPR)}, Long Beach, CA, USA, Jun. 2019, pp. 11028--11037.


\bibitem{Hui19:ACMMM}
Z. Hui, X. Gao, Y. Yang and X. Wang, ``Lightweight image super-resolution with information multi-distillation network,''
\emph{Proceedings of the 27th ACM International Conference on Multimedia (ACM MM)}, Nice, France, Oct. 2019, pp. 2024--2032.


\end{thebibliography}
\end{document}